\documentclass[letterpaper]{article} 
\usepackage{aaai23}  
\usepackage{times}  
\usepackage{helvet}  
\usepackage{courier}  
\usepackage[hyphens]{url}  
\usepackage{graphicx} 
\usepackage{natbib}  
\usepackage{caption} 
\usepackage{algorithm}
\usepackage{algorithmic}
\usepackage{amsfonts,amssymb} 
\usepackage{subfigure}
\usepackage{pgfplots}
\usepackage{dblfloatfix}
\usepackage{footnote}
\makesavenoteenv{tabular}
\makesavenoteenv{table}
\usepackage{newfloat}
\usepackage{listings}

\DeclareCaptionStyle{ruled}{labelfont=normalfont,labelsep=colon,strut=off} 
\floatstyle{ruled}
\newfloat{listing}{tb}{lst}{}
\floatname{listing}{Listing}
\title{PTab: Using the Pre-trained Language Model for Modeling Tabular Data}
\author{
    Guang Liu \textsuperscript{\rm 1}\equalcontrib, Jie Yang \textsuperscript{\rm 1 \rm 2} \equalcontrib, Ledell Wu \textsuperscript{\rm 1} \thanks{corresponding author}\\
}
\affiliations{
    \textsuperscript{\rm 1} Beijing Academy of Artificial Intelligence, Beijing, China\\
    \textsuperscript{\rm 2} Beijing Jiaotong University\\
    liuguang@baai.ac.cn, 21120149@bjtu.edu.cn, wuyu@baai.ac.cn

}

\usepackage{bibentry}

\begin{document}

\maketitle

\begin{abstract}
Tabular data is the foundation of the information age and has been extensively studied. Recent studies show that neural-based models are effective in learning contextual representation for tabular data. The learning of an effective contextual representation requires meaningful features and a large amount of data. However, current methods often fail to properly learn a contextual representation from the features without semantic information. In addition, it's intractable to enlarge the training set through mixed tabular datasets due to the difference between datasets. To address these problems, we propose a novel framework PTab, using the Pre-trained language model to model Tabular data. PTab learns a contextual representation of tabular data through a three-stages processing: Modality Transformation(MT), Masked-Language Fine-tuning(MF), and Classification Fine-tuning(CF). 
We initialize our model with a pre-trained Model (PTM) which contains semantic information learned from the large-scale language data. Consequently, the contextual representation can be learned effectively during the fine-tuning stages. In addition, we can naturally mix the textualized tabular data to enlarge the training set to further improve the representation learning.  We evaluate PTab on eight popular tabular classification datasets. Experimental results show that our method has achieved a better average AUC score in supervised settings compared to the state-of-the-art baselines(e.g. XGBoost), and outperforms counterpart methods under semi-supervised settings. We present visualization results that show PTab has well instance-based interpretability.
\end{abstract}

\section{Introduction}
Tabular data is essential for modern daily life and has been widely studied. It's the core component of many industrial applications, e.g., risk management of credit cards and insurance underwriting. In the last decades, researchers from various fields have studied effective ways of modeling tabular data. As far as we knew, the tree-based methods, e.g. XGBoost~\cite{chen2016xgboost}, have dominated this field due to their superior performance on feature extraction. 
\begin{figure}[htbp] 
\centering 
\includegraphics[width=0.46\textwidth]{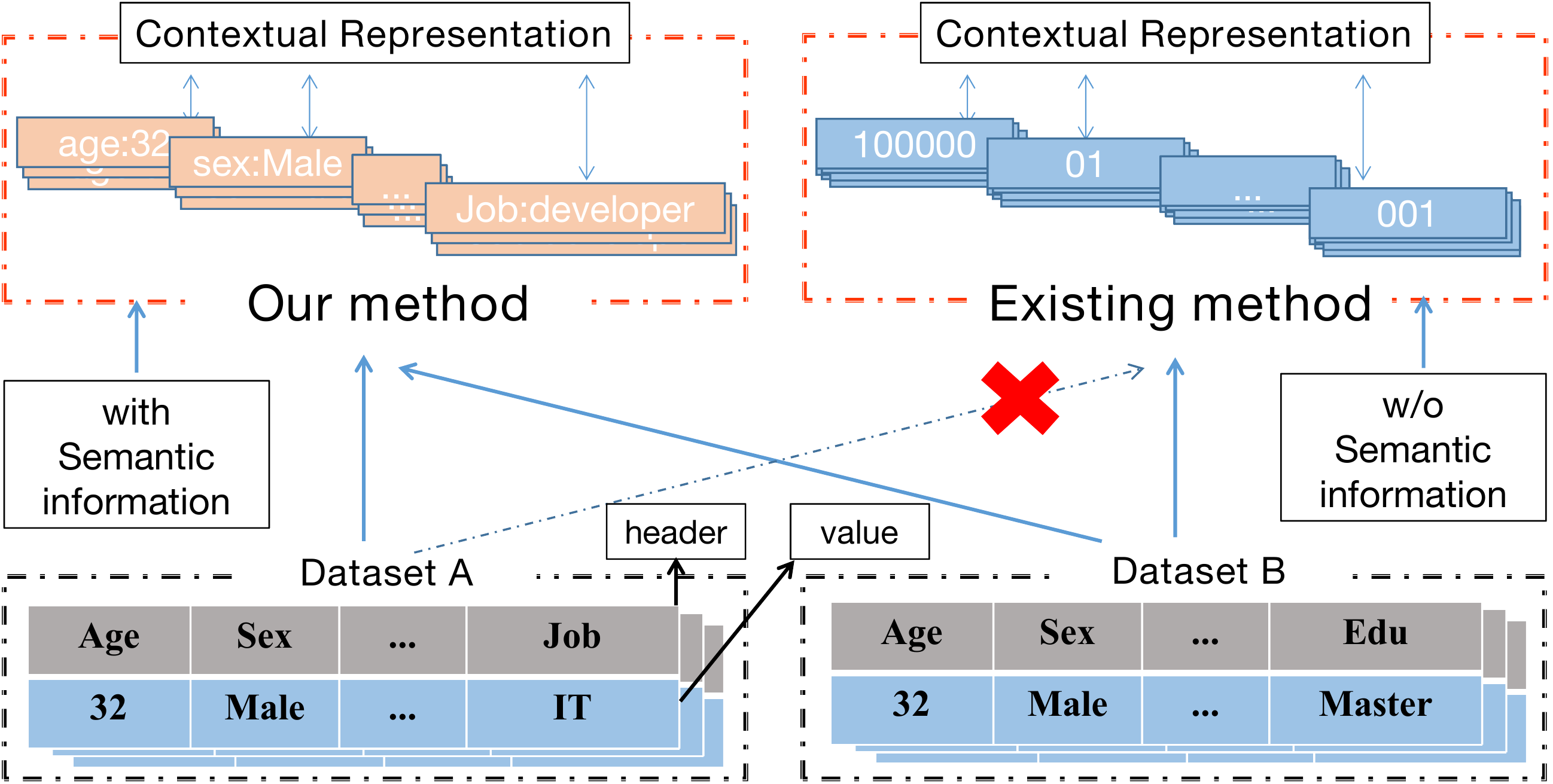} 
\caption{The illustration of methods on tabular modeling. The red folk means the mixing of datasets is not feasible. Each column in the examples is a field,  and it contains a header and its values.} 
\label{fig:fig1} 
\end{figure}

Recently, many neural-based methods have been proposed to learn contextual representations from tabular data~\cite{huang2020tabtransformer,yoon2020vime,arik2021tabnet}. Similar to the popular Masked Language Model (MLM) in Nature Language Processing~\cite{devlin2018bert}, TabTransfomer~\cite{huang2020tabtransformer} tries to learn a contextual representation by masking/corrupting the values of some random fields and predicting/recovering them based on the context. It uses a multi-layer transformer to handle the value of categorical fields and a Multi-Layer Perceptron(MLP) to handle the value of numerical fields. SAINT~\cite{somepalli2021saint} proposes projecting the value of numerical fields into an embedding, and 
uses a unified transformer to handle both numerical and categorical fields. In addition, it adopts multiple self-supervised learning tasks and data augmentation to further improve the contextual representation. These methods have achieved promising improvement compared to tree-based methods in semi-supervised settings. Intuitively, to lean an effective contextual representation requires a large amount of data and meaningful features. 

However, current methods often fail to properly learn a contextual representation from the features without semantic information. Besides, it's intractable to enlarge the training set through mixed tabular datasets due to the difference between datasets. Existing  methods, such as XGBoost~\cite{chen2016xgboost} and TabTransfomer~\cite{huang2020tabtransformer}, often lead to information loss during the feature transformation. As shown in Fig~\ref{fig:fig1}, Information technology (IT) is semantically close to engineers before the transformation but may be very distant after the transformation. Under such a condition, the current neural-based method often achieves under-performance compared to the tree-based methods~\cite{grinsztajn2022tree}. Additionally, the values of numeric fields tend to be in an ambiguous situation. For example, the value of age and shoe size both can be $32$, and they may be transformed into identical representations. The transformation process ignores the headers for the values which may help to improve the contextual representation. Without the information in the header, we need more data to learn a precise contextual representation. Noticeably, the mixing of different datasets to enlarge the size of the training set is not practical in the tabular domain. The difference in headers and values between different datasets makes it harder for the neural-based methods to learn a proper contextual representation. So, the technique that can model the tabular data with semantical information and mixed datasets is needed to be explored. 

\begin{figure*}[htbp]
\centering
\subfigure[Modality transformation]{
\includegraphics[width=1\columnwidth]{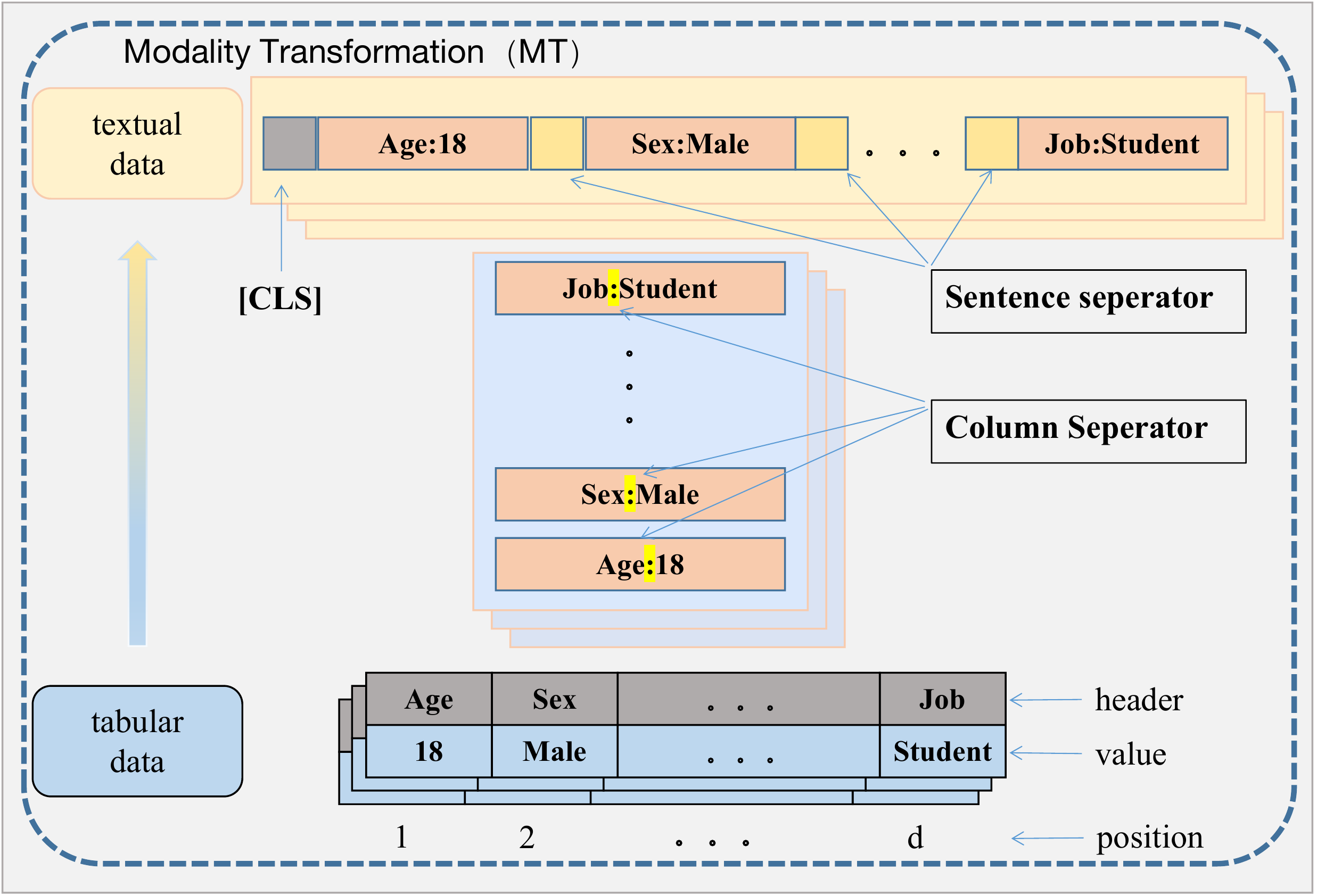}
\label{fig:fig2_1}
}
\subfigure[Training process]{
\includegraphics[width=1\columnwidth]{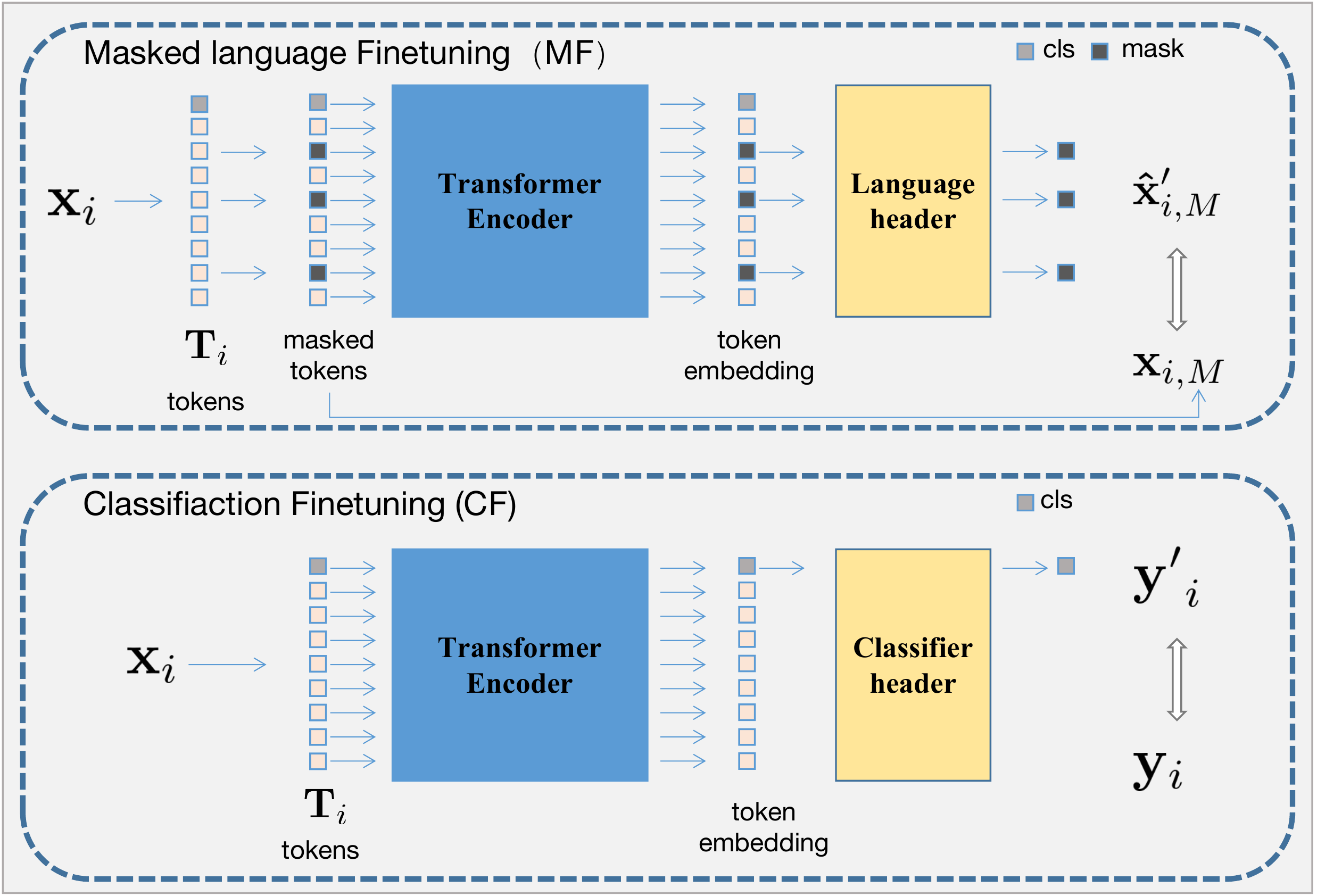}
\label{fig:fig2_2}
}
\caption{The framework of our proposed method}
\label{fig:fig2}
\end{figure*}

To address these problems,  we propose a novel framework, using the Pre-trained language model to model Tabular data (PTab), to enhance the contextual representation of tabular data with semantic information and makes the training on mixed datasets feasible. PTab consists of three stages: Modality Transformation(MT), Masked-language fine-tuning(MF) and Classification fine-tuning(CF). MT stage is fundamental to the framework. It textualizes the tabular data to enrich its semantical feature and diminish the difference between different tabular datasets. To this end, it uses a simple lexicon and an ordered syntax to describe the headers and values in a sample of tabular data. The information that describes the meanings of the field can be used easily, e.g., the header information of each fields. Consequently, we can utilize the knowledge in the PTM to enhance the learned representation through the MF stage. CF stage adjusts the learned contextual representation for the downstream task. It's worth mentioning that textualization diminishes the difference between datasets. By doing so, the model training on mixed textualized tabular datasets can be realized smoothly. The tabular datasets from the similar domains can be used to improve representation learning for the corresponding downstream task. The major contributions of this paper are summarized as follows:
\begin{itemize}
    \item We propose a novel framework PTab, using the Pre-trained language model to model Tabular data. PTab can improve the contextual representation learned from tabular data by incorperate extra semantical information.
   \item We evaluate PTab on eight binary classification datasets. The extensive experimental results shows that PTab outperforms the state-of-the-art baselines, including strong tree-based baseline XGBoost and GBDT, in terms of average AUC score in both supervised/semi-supervised settings. 
   \item We firstly propose a feasible way for training on mixed tabular datasets. The results of mixed dataset training demonstrate that PTab can get higher classification performance by training on mixed textualized tabular datasets in similar domains, and vice versa.
    \item  Interestingly, the visualization results demonstrate that PTab trained model has well instance-based interpretability for feature importance and feature semantical similarities, while the tree-based method is hard to achieve.
\end{itemize}

\section{Related works}





Tabular data modeling is the fundamental task in financial and insurance domains. Many researchers focus on extracting features and building models from tabular data~\cite{gorishniy2021revisiting,borisov2021deep,shwartz2022tabular,gorishniy2021revisiting}. These studies can be divided into numerical-based methods and text-based methods.

Numerical-based methods mainly focus on using neural-based methods to learn contextual representations from numerical tabular data~\cite{gorishniy2021revisiting,shwartz2022tabular,yoon2020vime}. TabNet~\cite{arik2021tabnet} firstly proposes using self-supervised learning to learn the contextual representation. SAINT~\cite{somepalli2021saint} uses data augmentation and multi-task learning to enhance the learning process. Subtab~\cite{ucar2021subtab} turns the task of learning from tabular data into a multi-view representation learning problem by dividing the input features into multiple subsets. This line of research usually uses contrastive learning to learn a representation from unlabeled numerical data and then fine-tuned on labeled numerical data. However, such methods often suffer from a shortage of labeled data~\cite{xu2019modeling}. Without enough labeled data, it's intractable to learn an effective representation.

Textual-based methods mainly focus on learning table representation for table-related tasks~\cite{iida2021tabbie,deng2022turl,chen2019learning}. Different from numerical-based methods, textual-based method tries to encode multi-rows of tabular data into one representation. Table2vec~\cite{zhang2019table2vec}  considers different table elements, such as caption, field headers and values, for training word and entity embeddings. Tabert~\cite{yin2020tabert} jointly learns representations for natural language sentences and semi-structured tables. TURL~\cite{deng2022turl} captures factual knowledge in relational tables. TUTA~\cite{wang2021tuta} designs a bi-dimensional tree to define value coordinates and value distance in generally structured tables. These methods mainly focus on specific downstream tasks in Natural Language Processing (NLP).  Few works have explored the way to model one row of tabular data as numerical-based methods.

Accordingly, both numerical- and textual-based methods learn contextual representation contrastively: mask/corrupt some elements in a sample and predict/recover them by their context. For example, TabTransfomer~\cite{huang2020tabtransformer}, TabNet\cite{arik2021tabnet}, and SAINT\cite{somepalli2021saint}  all learns tabular representation by applying the Masked Language Model (MLM) to categorical fields. Tabbie~\cite{iida2021tabbie} and TUTA \cite{wang2021tuta} also use MLM for textual tabular data representation learning. And MLM-like method works well on textual tabular data and often not on numerical tabular.

\begin{table*}[b]
\caption{The statistics of all dataset.\label{tab:tab1}}
\centering
\setlength{\tabcolsep}{2.5mm}{
\begin{tabular}{cccccccccc}
\hline
Dataset           & BM & AD & IN & OS & ST & BC  & SB & QB \\\hline
Num of examples        & 45211          & 34190  & 32561       & 12330           & 10000    & 7043        & 2583         & 1055     \\
Num of Columns          & 16             & 24     & 14          & 17              & 11       & 20         & 18           & 41       \\
Num of Textual  Category$^1$      &  8               &   0    &   8          &  3               & 3          &    16       &     4          &0 \\
Domain          &   business              & income      &   income          &   shopping              &   business        &   business       &  geology            &chemistry \\               
Posivie ratio(\%)     & 11.7           & 85.4   & 24.1        & 15.5            & 20.4     & 26.5            & 6.6          & 33.7 \\\hline   
\end{tabular}}
$^1$The textual category means the value of the feature is a textual phrase or sentence.
\end{table*}

\section{Task formulation}
For a given $n$ samples tabular dataset, $\{\textbf{X}_i, y_i\},\textbf{X}_i\in \mathbb{R}^{D}, i\in[1,n]$, we formulate the tabular data modeling as follow,
\begin{equation}
    f_\theta(\textbf{X}_i;\textbf{H}) = y_i\,.
\end{equation}
Here, $f_\theta$ is a model with parameter $\theta$, the $i_{th}$ row of tabular data $\textbf{X}_i$ contains $D$ columns, $\textbf{X}_i \sim \{x_0, x_1,\cdots, x_D\}$,$\textbf{H} \sim \{h_0,h_1,\cdots,h_D\}$ represent the shared header descriptions for all $\textbf{X}$, $y_i$ is the category label. Noticeable, the $H$ provides extra information for tabular data modeling.

\section{Methodology}
The proposed PTab is a three-stage framework to model textualized tabular data with the pre-trained model, as illustrated in Fig.~\ref{fig:fig2}. The first stage is Modality Transformation (MT) which textualizes the tabular data with a simple lexicon and ordered syntax. The second stage is Masked language fine-tuning(MF) which performs a randomly masking and predicting process to learn a contextual representation from textualized tabular data. The third stage is Classification fine-tuning(CF) which adjusts the learned contextual representation for a specific task. 
\subsection{Modality transformation}
As shown in Fig~\ref{fig:fig2_1}, the MT stage textualizes the tabular dataset with a simple lexicon and ordered syntax. This is a simple and effective way to pull different tables into the same semantic space for representation learning~\cite{deng2022turl}. This transformation has two major steps, phrase and sentence construction.

For phrase construction, we simply join the header $h_i$ and value $x_i$ of $i_{th}$ column with a column separator $:$ as a phrase $p_i=\{h_i:x_i\}$.  For sentence construction, we join the phrases with sentence separator $[SEP]$ by their appearance order in the row. In the end, we get the sentence $S_i=\{p_0,p_1,\cdots,p_d\}$ for the tabular data $\{X_i,H\}$. Recently, textual-based method unified-SKG~\cite{xie2022unifiedskg} uses " " to as column seperator and  $|$ as columns separator. We find out that the choice of column separator makes an insignificant difference in the final performance.

\subsection{Masked language fine-tuning}
In the MF stage, the sentence $S$ of textualized $X$ is fed into a pre-trained model for contextual representation learning. For example, the tabular data $X=\{32, Male, driver\}$ with headers $H=\{Age, Sex, Job\}$ can be transformed into a textual sentence. So, the $X$ is  translated into a sentence $S=\{Age:32[SEP]Sex:Male[SEP]Job:driver\}$. Before contextual learning, we tokenize the sentence $S_i$ into tokens $T = \{t_0,t_1,\cdot,t_k\}$.  For the final task tuning, we add the $[CLS]$ token at the beginning of each sentence. Following, we randomly mask a certain ratio of tokens with $[MASK]$. Then, we train the model to predict the masked token. The learning target can be formalized as follows,
\begin{equation}
    L_{mf}(X_M|X_{-M})=\frac{1}{K}\sum_{k=1}^{K}\log p(x_{m_i}|X_{-M};\theta)\,.
\end{equation}
Here, $M=\{m_0,m_1,\cdots,m_K\}$ denotes the masked tokens indexes in sentence $X$, $K$ is the number of masked tokens. Similar to BERT~\cite{devlin2018bert}, the MF stage aims to predict the masked token depend on corresponding contents. By doing so, the contextual representation can be properly learned. 

\noindent\textbf{Training on mixed datasets}
Noticeably, the MT stage can be performed on mixed textualized tabular datasets.  Different from the current methods in modeling tabular data, we perform contextual representation learning on textualized tabular data. Similar to textual-based methods, different datasets can be fed into a PTM for contextual representation learning~\cite{yin2020tabert}. In textual modality, the difference between tabular datasets degenerates as the difference between sentences. We can treat the mix of data from the different datasets as the mix of sentences from different domains. Intuitively, the sentence from the same domain will benefit the representation learning(see Section~\ref{sec:mixed}).

\subsection{Classification fine-tuning}
In CF stage, we  fine-tuning the model on classification tasks. This stage can adjust the learned contextual representation for the specific task. The model is initialized by the model with the lowest loss in MF stage, following the setting in previous work~\cite{hu2019learning}.  We use the embedding of $[CLS]$ token to represent the sentence. On top of the sentence representation, we add a randomly initialized linear projection layer. The output of projection layer is the final predictions.  The target of CF stage is to learn a function from $S_i$ to $y_i$ as follows,
\begin{equation}
    f(S_i)_\theta' = y_i\,.
\end{equation}
Here, the learning target is a cross-entropy loss on two classes $1/0$, the model $f_\theta'$ has different linear projection layers with $f_\theta$. The semantic of the sentence representation will be changed gradually towards the target of the task.

\section{Experimental settings}
\begin{table*}[htbp]
\centering
\caption{The results on full-size dataset. \label{tab:tab2}}
 \setlength{\tabcolsep}{1.3mm}{
\begin{tabular}{ccccccccc|c}\hline
\centering
Dataset    & BM  & AD             & IN      & OS   & ST            & BC           & SB        & QB   & AVG        \\\hline
TabNet*    & 0.885\tiny{±0.017}     & 0.663\tiny{±0.016}        & 0.875\tiny{±0.006}       & 0.888\tiny{±0.020}        & 0.785\tiny{±0.024}         & 0.816\tiny{±0.014}         & 0.701\tiny{±0.051}         & 0.860\tiny{±0.038}          &0.809\tiny{±0.087}  \\
LR*        & 0.911\tiny{±0.005}     & 0.721\tiny{±0.010}        & 0.899\tiny{±0.002}       & 0.908\tiny{±0.015}       & 0.828\tiny{±0.013}         & 0.844\tiny{±0.01}          & 0.749\tiny{±0.068}         & 0.847\tiny{±0.037}  &0.838\tiny{±0.071}        \\

TabTransfomer*  & 0.934\tiny{±0.004}     & 0.737\tiny{±0.009}        & 0.906\tiny{±0.003}       & 0.927\tiny{±0.010}        & 0.856\tiny{±0.005}         & 0.835\tiny{±0.014}         & 0.751\tiny{±0.096}         & 0.918\tiny{±0.038} &0.858\tiny{±0.078}          \\

GBDT*      & 0.933\tiny{±0.003}     & \textbf{0.756\tiny{±0.011}}        & 0.906\tiny{±0.002}       & \textbf{0.930\tiny{±0.008} }       & \textbf{0.859\tiny{±0.009}}         & \textbf{0.847\tiny{±0.016}}         & 0.756\tiny{±0.084}         & 0.913\tiny{±0.031}   &0.863\tiny{±0.073}        \\

\hline
TabNet@    & 0.921\tiny{±0.003} & 0.711\tiny{±0.011} & 0.906\tiny{±0.004} & 0.913\tiny{±0.006} & 0.841\tiny{±0.013} & 0.826\tiny{±0.010} & 0.701\tiny{±0.018} & 0.643\tiny{±0.053} & 0.808\tiny{±0.077} \\
LR@        & 0.889\tiny{±0.003}     & 0.656\tiny{±0.008}        & 0.848\tiny{±0.006}       & 0.896\tiny{±0.006}       & 0.745\tiny{±0.008}         & 0.845\tiny{±0.006} & 0.697\tiny{±0.044} & \textbf{0.921\tiny{±0.025}}&0.812\tiny{±0.099}   \\

SAINT@     & 0.810\tiny{±0.012}    & 0.728\tiny{±0.004}        & 0.909\tiny{±0.003}      & 0.907\tiny{±0.009}      & 0.814\tiny{±0.023}        & 0.814\tiny{±0.008}        & 0.766\tiny{±0.030}          & 0.918\tiny{±0.002}        &0.833\tiny{±0.071}  \\
XGBoost@   & 0.933\tiny{±0.002}     & 0.741\tiny{±0.017}        & 0.919\tiny{±0.003}       & \textbf{0.930\tiny{±0.002}}       & 0.854\tiny{±0.012}         & 0.842\tiny{±0.004}         & 0.753\tiny{±0.026}         & 0.914\tiny{±0.040}        &0.861\tiny{±0.078}   \\
Our        & \textbf{0.939\tiny{±0.005 }  }       & 0.730\tiny{±0.008}         & \textbf{0.926\tiny{±0.002} }           & \textbf{0.930\tiny{±0.003}}       & \textbf{0.859\tiny{±0.008}}         & \textbf{0.847\tiny{±0.002} }        & \textbf{0.783\tiny{±0.014}}         & 0.916\tiny{±0.022}       &\textbf{0.866\tiny{±0.077} }  \\\hline
\end{tabular}}
 $*$ means cited from TabTransfomer~\cite{huang2020tabtransformer}. @ means our reproduced results.
\end{table*}
\subsection{Data}
As shown in Tab.~\ref{tab:tab1}, we evaluate our method on eight classification datasets, which are used in recent work~\cite{huang2020tabtransformer}. Bank Marketing \textbf{(BM)} is related to direct marketing campaigns (phone calls) of a Portuguese banking institution. 1995 INcome~\textbf{(IN)} aims to determine whether a person makes over 50K a year. Online Shoppers \textbf{(OS)} aim to detect users who browse products online will eventually choose to buy. Seismic Bumps \textbf{(SB)} describe the problem of high energy seismic bumps forecasting in a coal mine. ShruTime \textbf{(ST)} contains details of a bank's customers and the target variable is a binary variable reflecting the fact whether the customer left the bank (closed his account) or he continues to be a customer. Qsar Bio \textbf{(QB)} aims to classify $1055$ chemicals into $2$ classes (ready and not ready biodegradable). BlastChar \textbf{(BC)} aims to predict behavior to retain customers. These datasets are collected from different domains. \textbf{BM}, \textbf{ST}, and \textbf{BC} are from the business domain. \textbf{SB} is from the mining industry. \textbf{BC} is from the field of chemistry. Besides, the data have different ratios of category features. Most datasets have both numeric features and category features, except \textbf{QB} and \textbf{AD}. We evaluate our method on each dataset under five-fold cross-validation. In each fold, the split ratio is 65/15/20\%, following the settings in TabTransfomer for a fair comparison.


\subsection{Baseline models}
We compare our method with six baseline models: 1)logistic regression model (LR), the tree-based model 2)XGBoost~\cite{chen2016xgboost} and 3)GBDT~\cite{friedman2001greedy}’, the neural-based model 4)TabNet~\cite{arik2021tabnet}, 5) TabTransfomer~\cite{huang2020tabtransformer},and 6)SAINT~\cite{somepalli2021saint}. For fair comparison, we empirically tune the key hyperparameters for the baseline model seperately and choose the best resutls. Specially, we apply a greed search for XGBoost on number of trees $\in [500,600,\cdots,1000] $,  depth $\in [4,8,16]$ and learning rate $ \in [0.05,0.1,0.15,0.2]$.   

\subsection{Settings}
In Modality Transformation (MT) stage, we simplify the header information and category values to avoid exceeding the maximun length of $512$. The data are textualized, tokenized and padded into the same length. In Masked language Fintuning(MF) stage, we set the mask ratio to $15\%$, and train the model $10$ epochs with batch size $16$ and learning rate $4e^{-5}$. We evaluate the model on validation set at the end of each epoch and select the best model as the initial weights for Classification fine-tuning(CF). In CF stage, we train our model for $40$ epoches with batch size $16$ and learning rate $2e^{-5}$. The one with the best auc score on validation set is selected for the final testing. We report the mean and variation of AUC score in five-fold cross-validation. We choose the base version BERT~\cite{devlin2018bert} as our initial model which has $12$ attention layers, $12$ attention heads and $768$ hidden state size. In MF stage, the language head of BERT is added on top of the model to get  the final vocabulary size predictions for contextual representation learning. In CF stage, we replace the language head with a random initialized classification head to adjust the $[CLS]$ embedding for better adaption to classification. In both supervised and semi-supervised settings, we use a linear
learning rate warmup~\cite{he2016deep} in MF stage. 

\section{Results and analysis}
\subsection{Supervised learning}
To evaluate our proposed method of contextual representation learning, we compare our model with six baseline models. As shown in Tab~\ref{tab:tab2}, our model outperforms all baselines in terms of average AUC scores. For example, we achieve a 4\% improvement in average score compared to SAINT. Different from SAINT, our model uses a pre-trained model as an initial model and uses textualized tabular data. These two major difference brings extra semantic information for our model to learn a more precise contextual representation. 

Notably, our model outperforms the state-of-the-art tree-based methods, XGBoost and GBDT, on six out of eight datasets. The neural-based model is often under-performance than the tree-based model due to the sensitivity to uninformative features~\cite{grinsztajn2022tree}. The superior performance of our model may come from the textual modality transformation and attention mechanism, which help to enrich the features information and selectively ignore the uninformative features.

Besides, our method achieves significantly higher AUC results on datasets with textual category features, such as \textbf{BM/IN/SB}. The textual category features mean the feature value is a phrase or sentence. Tab~\ref{tab:tab1} shows the number of textual category features and total features on all datasets. And our method achieves comparable results on numerical features only datasets, such as \textbf{AD/QB}. Such results indicate that our method benefits from the pre-training process which can extract information from meaningful features.


\begin{table}[bp]
\setlength{\tabcolsep}{1.7mm}{
\caption{AUC score for semi-supervised learning models on all datasets with 50 data points. Values are the mean over 5 cross-validation splits.\label{tab:tab3}}
\begin{tabular}{ccc|cc}
\hline
Dataset & TabTrans  & GBDT(PL)    & SAINT       & Our         \\\hline
BM      & 0.735\tiny{±0.040}  & 0.688\tiny{±0.057} & 0.677\tiny{±0.074} & \textbf{0.769\tiny{±0.008}} \\
AD      & 0.613\tiny{±0.014}   & 0.519\tiny{±0.024} & 0.543\tiny{±0.023} & \textbf{0.617\tiny{±0.021}} \\
IN      & \textbf{0.862\tiny{±0.018}}   & 0.685\tiny{±0.084} & 0.685\tiny{±0.08}  & 0.828\tiny{±0.006} \\
OS      & 0.780\tiny{±0.024}   & \textbf{0.818\tiny{±0.032}} & 0.807\tiny{±0.023} & 0.755\tiny{±0.018} \\
ST      & \textbf{0.741\tiny{±0.019}}   & 0.651\tiny{±0.093} & 0.662\tiny{±0.034} & 0.722\tiny{±0.025} \\
BC      & \textbf{0.822\tiny{±0.009}}  & 0.729\tiny{±0.053} & 0.757\tiny{±0.021} & 0.787\tiny{±0.041} \\
SB      & 0.738\tiny{±0.068}    & 0.601\tiny{±0.071} & 0.720\tiny{±0.029} & \textbf{0.749\tiny{±0.046}} \\
QB      & 0.869\tiny{±0.036}   & 0.804\tiny{±0.057} & 0.861\tiny{±0.031} & \textbf{0.876\tiny{±0.034}}                   \\ \hline
AVG & \textbf{0.770\tiny{±0.084}} & 0.687\tiny{±0.100} & 0.714\tiny{±0.097} & \textit{0.763\tiny{±0.076}} \\
\hline
\end{tabular}}

\end{table}

\subsection{Semi-supervised Learning}
To further evaluate the ability of PTab on learning contextual representation, we conduct three groups of experiments in semi-supervised settings.  Specifically, we sample a fixed amount of labeled examples from the training set and remove the labels of the unselected training examples. In this way, we set up three groups of experiments with $50$, $200$, and $500$ labeled examples, which is a commonly used settings~\cite{huang2020tabtransformer,wei2019eda,sohn2020fixmatch}. Five-fold cross-validation is performed for each experiment and the average and standard deviation of the AUC score are reported. We choose the state-of-the-art models on semi-supervised settings, SAINT and TabTransfomer, for comparison. We choose the popular tree-based method, XGBoost, plus Pseudo labeling as the baseline.

\begin{table}[htbp]
\setlength{\tabcolsep}{1.7mm}{
\caption{AUC score for semi-supervised learning models on all datasets with 200 fifine-tune data points. Values are the mean over 5 cross-validation splits.\label{tab:tab4}}
\begin{tabular}{ccc|cc}
\hline
    Dataset        &  TabTrans & GBDT(PL)    & SAINT       & Our         \\\hline
    BM  & 0.838\tiny{±0.010}      & 0.802\tiny{±0.012} & 0.835\tiny{±0.017} & \textbf{0.856\tiny{±0.030}} \\
    AD          & 0.614\tiny{±0.012}   & 0.572\tiny{±0.04}  & 0.587\tiny{±0.016} & \textbf{0.631\tiny{±0.016}} \\
    IN     & \textbf{0.875\tiny{±0.011}}     & 0.822\tiny{±0.02}  & 0.814\tiny{±0.04}  & 0.862\tiny{±0.004} \\
    OS & 0.838\tiny{±0.024}    & 0.846\tiny{±0.019} & 0.869\tiny{±0.010} & \textbf{0.897\tiny{±0.011}} \\
    ST       & \textbf{0.783\tiny{±0.024}}      & 0.750\tiny{±0.050}   & 0.643\tiny{±0.048} & 0.782\tiny{±0.026} \\
    BC      & \textbf{0.841\tiny{±0.014}}    & 0.783\tiny{±0.017} & 0.804\tiny{±0.006} & 0.812\tiny{±0.016} \\
    SB   & 0.708\tiny{±0.083}   & 0.603\tiny{±0.023} & 0.758\tiny{±0.022} & \textbf{0.761\tiny{±0.013}} \\
    QB        & \textbf{0.889\tiny{±0.030}}        & 0.855\tiny{±0.035} & 0.885\tiny{±0.023} & \textbf{0.889\tiny{±0.032}} \\\hline
    AVG & \textit{0.798\tiny{±0.094}}  & 0.754\tiny{±0.108} & 0.774\tiny{±0.107} & \textbf{0.811\tiny{±0.088}} \\\hline
\end{tabular}}

\end{table}

\begin{table}[htbp]
 \setlength{\tabcolsep}{1.7mm}{
 \caption{AUC score for semi-supervised learning models on all datasets with 500 fifine-tune data points. Values are the mean over 5 cross-validation splits.\label{tab:tab5}}
\begin{tabular}{ccc|cc}
\hline
Dataset        & TabTrans  & GBDT(PL)    & SAINT       & Our         \\\hline
BM  & 0.860\tiny{±0.016}  & 0.838\tiny{±0.019} & 0.853\tiny{±0.025} & \textbf{0.884\tiny{±0.003}} \\
AD          & 0.647\tiny{±0.008}    & 0.647\tiny{±0.030} & 0.651\tiny{±0.024} & \textbf{0.672\tiny{±0.032}} \\
IN     & \textbf{0.880\tiny{±0.007}}        & 0.839\tiny{±0.013} & 0.875\tiny{±0.018} & 0.875\tiny{±0.006} \\
OS & 0.861\tiny{±0.014}         & 0.865\tiny{±0.011} & 0.895\tiny{±0.008} &\textbf{ 0.918\tiny{±0.008}} \\
ST       & 0.815\tiny{±0.004}    & 0.788\tiny{±0.019} & 0.772\tiny{±0.020} & \textbf{0.834\tiny{±0.031}} \\
BC      & \textbf{0.839\tiny{±0.015}}  & 0.795\tiny{±0.021} & 0.815\tiny{±0.005} & 0.836\tiny{±0.011} \\
SB   & 0.729\tiny{±0.069}   & 0.666\tiny{±0.063} & 0.760\tiny{±0.025} & \textbf{0.766\tiny{±0.021}} \\
QB        & 0.889\tiny{±0.038}  & \textbf{0.908\tiny{±0.024}} & 0.896\tiny{±0.030} & 0.902\tiny{±0.030}:                        \\ \hline
AVG & \textit{0.815\tiny{±0.085}} & 0.793\tiny{±0.093} & 0.815\tiny{±0.084} & \textbf{0.836\tiny{±0.082}} \\\hline
\end{tabular}}

\end{table}

\begin{table*}[bhtp]
\setlength{\tabcolsep}{2.8mm}{
\caption{Results of ablation study}
\label{tab:tab6}
\begin{tabular}{llllllllllll}\hline
Dataset        & BM & AD       & IN   & OS & ST      & BC     & SB & QB    \\\hline
our            & 0.933\tiny{±0.012}         & 0.730\tiny{±0.008}   & 0.926\tiny{±0.002}     & 0.930\tiny{±0.003}     & 0.859\tiny{±0.008}   & 0.846\tiny{±0.004}    & 0.783\tiny{±0.014}  & 0.916\tiny{±0.022}   \\
 \quad$-sep$ & 0.886\tiny{±0.024}    & 0.674\tiny{±0.021}  & 0.915\tiny{±0.016}   & 0.902\tiny{±0.007}     & 0.848\tiny{±0.018}   & 0.826\tiny{±0.011}   & 0.754\tiny{±0.017}  & 0.875\tiny{±0.012}   \\
\quad$-mlm$     & 0.782\tiny{±0.061}    & 0.596\tiny{±0.044} & 0.912\tiny{±0.009} & 0.863\tiny{±0.034}  & 0.843\tiny{±0.009} & 0.839\tiny{±0.006}  & 0.750\tiny{±0.022} & 0.838\tiny{±0.043} \\
\hline
\end{tabular}}
\end{table*}

Tab~\ref{tab:tab3} show results on $50$ labeled examples. The experiment results show that our method outperforms SAINT with a significant margin in terms of average AUC score. In addition, our method achieves the best AUC score on half of the eight datasets. Our methods also achieve the second-best average AUC score. We think that the under-performance may be caused by high variance under different random seeds when the labeled data is few. 

As illustrated in Tab~\ref{tab:tab4} and ~\ref{tab:tab5}, our method achieves the highest AUC score on 5 out 8 datasets and the highest average AUC score. The experimental results show that our method outperforms counterpart methods by a significant margin with more than 50 labeled examples in semi-supervised settings. For example, our method achieves a $0.836$ average AUC score which is 2.6\% higher than TabTransfomer/SAINT and 5.4\% higher than GBDT(PL).  Besides, our method achieves stable improvements in AUC score with the increase of labeled examples, while TabTransfomer(200) achieves lower AUC than TabTransfomer(50) on SB, and SAINT(200) achieves lower AUC than SAINT(50) on ST.

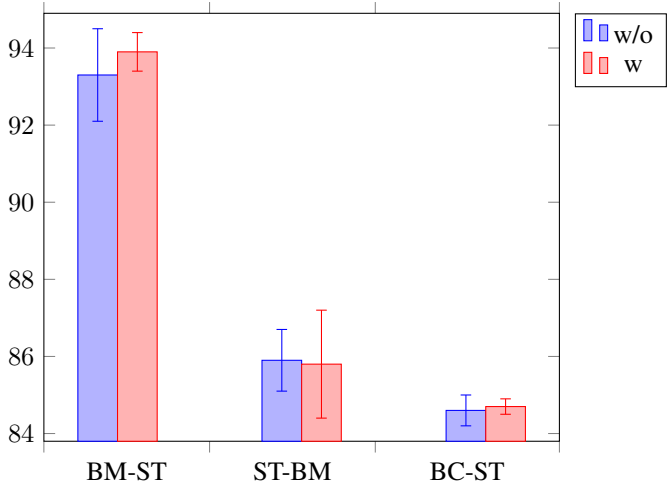
\begin{figure}[htbp]
\begin{tikzpicture}
\begin{axis}[
legend pos=outer north east,
enlargelimits={abs=0.4},
ybar=0pt,
bar width=15,
xtick={0.6,1.5, 2.4, 3.4},
xticklabels={BM-ST,ST-BM,BC-ST},
x tick label as interval
]

\addplot+[error bars/.cd,
y dir=both,y explicit]
coordinates {
    (1,93.3) +- (0.0, 1.2)
    (2,85.9) +- (0.0, 0.8)
    (3,84.6) +- (0.0, 0.4)
};
\addplot+[error bars/.cd,
y dir=both,y explicit]
coordinates {
    (1,93.9) +- (0.0, 0.5)
    (2,85.8) +- (0.0, 1.4)
    (3,84.7) +- (0.0, 0.2)
};
\legend{w/o,w}
\end{axis}
\end{tikzpicture}
\caption{Resutls of training on mixed in-domain datasets.}
\label{fig:fig3}
\end{figure}

\begin{figure}[htbp]
\begin{tikzpicture}
\begin{axis}[
legend pos=outer north east,
enlargelimits={abs=0.4},
ybar=0pt,
bar width=15,
xtick={0.6,1.5, 2.4,3.4},
xticklabels={BM-OS,ST-OS,BC-OS},
x tick label as interval
]

\addplot+[error bars/.cd,
y dir=both,y explicit]
coordinates {
    (1,93.3) +- (0.0, 1.2)
    (2,85.9) +- (0.0, 0.8)
    (3,84.6) +- (0.0, 0.4)
};
\addplot+[error bars/.cd,
y dir=both,y explicit]
coordinates {
    (1,90.6) +- (0.0, 2.4)
    (2,83.5) +- (0.0, 0.9)
    (3,81.6) +- (0.0, 0.13)
};

\legend{w/o,w}
\end{axis}
\end{tikzpicture}
\caption{Resutls of training on mixed out-of-domain datasets.}
\label{fig:fig4}
\end{figure}
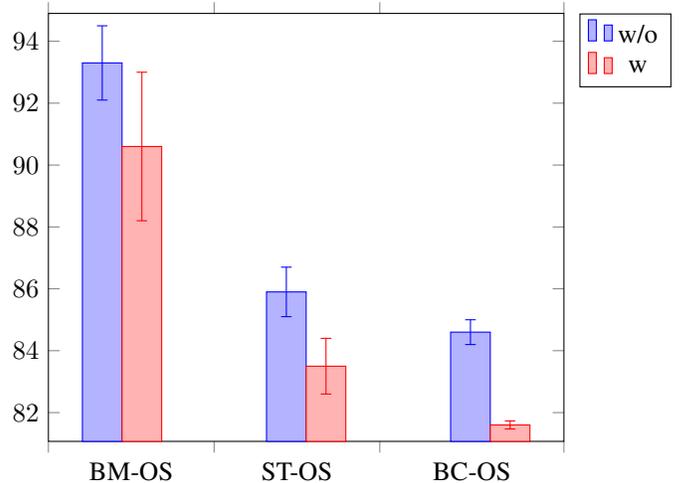



\subsection{Mixed dataset pre-training}
\label{sec:mixed}


To evaluate our proposed method for mixed datasets training, we set up groups of experiments. Mixing datasets in the similar domain is a commonly used method to enlarge the training dataset~\cite{li2021unified,li2022blip}. Existing models are specifically designed for a single dataset and are not extensible to others. Our method can tackle this problem by unsupervised pre-training on related task data and fine-tuning on target task data.

1) We train our model on mixed dataset (\textbf{BM},\textbf{ST} and \textbf{BC}) from the business domain, which is mentioned in the dataset description in Tab~\ref{tab:tab1}.  As shown in Fig~\ref{fig:fig3}, training on mixed in-domain datasets will help to improve the performance of tabular data modeling. For detail, \textbf{BM} aims to predict if the client will subscribe (yes/no) to a term deposit. \textbf{ST} aims to predict the fact whether the customer left the bank (closed his account) or continues to be a customer. \textbf{BC} aims to predict behavior to retain customers. Besides, they have some feature overlaps, e.g., \textbf{BC} and \textbf{ST} both contain the time of service ordered and the sex of custom. There is some overlap in feature names and values between these datasets. By textualizing tabular data, our model can learn correlations between features through attention mechanisms. The experimental results show that our model can benefit from the mixed in-domain datasets.

2)For comparison, we also train our model on mixed out-of-domain datasets, as illustrated in Fig.~\ref{fig:fig4}. The unrelated task data will significantly reduce the performance of the target task. For example, the \textbf{BM} and \textbf{OS} has feature overlap,e.g., education and job, but very different task definition. In this situation, the training on mixed datasets may lead to performance degradation.

\subsection{Ablation Study}


To analyze the effects of each stage in our proposed PTab,  we performed ablation experiments on all 8 datasets. As shown in Tab~\ref{tab:tab6}, $-sep$ represents the results of removal of sentence separator in the MT stage, and $-mlm$ indicates the results of removal of MF stage. We cannot apply classification without the CF stage, so we do not test the performance with the removal of this component. $-sep$ and $-mlm$ both lead to a decrease in performance significantly. Without the sentence separator, we get significant performance degradation on all datasets. That indicates the MT stage plays an important role in our methods. Moreover, the removal of the MF stage decreases the performance of our model further. Specifically, $-mlm$ has lower performance than $-sep$. The experiments show that the MT stage the is the key component in our proposed framework.

\begin{figure}[htbp] 
\centering 
\includegraphics[width=0.45\textwidth]{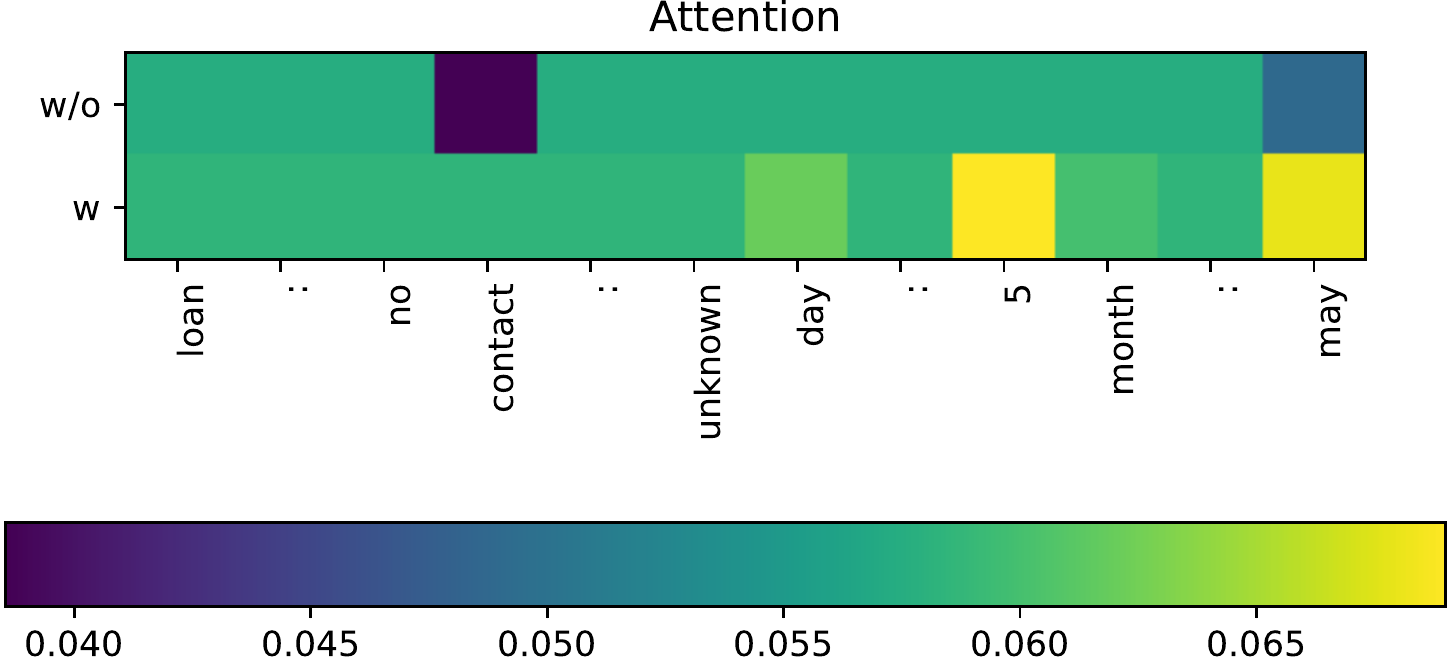} 
\caption{Visualization of attention weights for selected tokens of a randomly picked example from BM.} 
\label{fig:fig7} 
\end{figure}

\begin{figure}[htbp]
\centering
\subfigure[w/o Masked language Fine-tuning(MF)]{
\includegraphics[width=0.95\columnwidth]{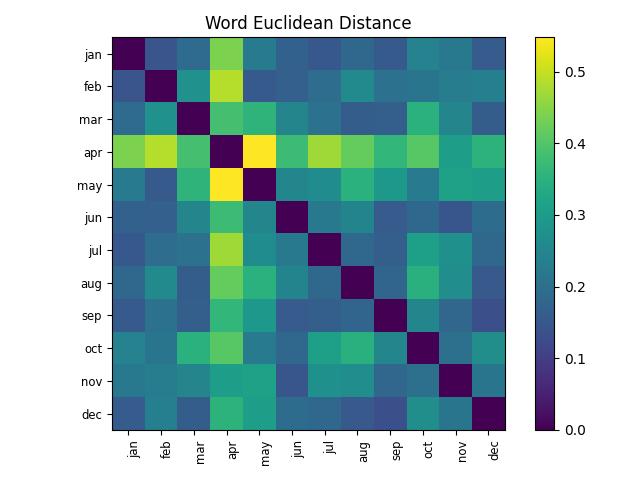}
\label{fig:fig6_1}
}
\subfigure[w/ Masked language Fine-tuning(MF)]{
\includegraphics[width=0.95\columnwidth]{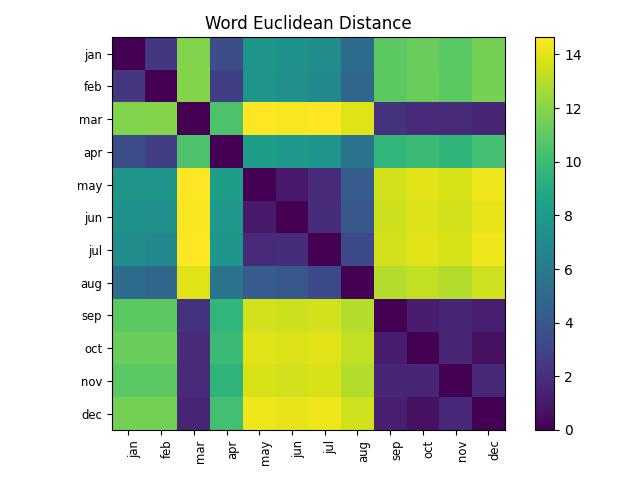}
\label{fig:fig6_2}
}
\caption{The Euclidean Distance of examples with different month value.}
\label{fig:fig6}
\end{figure}


\subsection{Visualization of Attention weights}
We visualize the attention score to analyze whether the attention scores can learn effective features from textualized tabular data. The  $[CLS]$ token is used for final classification. So the attention score of $[CLS]$ token reflect the importance of each features. The visualization of attention scores can help us to explain instant-based feature importance. We draw the attention scores of the $[CLS]$ of a random select example from BM dataset. The attention scores have changed significantly after fine-tuning. As shown in Fig.~\ref{fig:fig7},  $w/o$ denotes the attention scores of $[CLS]$ token without Masked language Fine-tuning(MF), $w$ denotes the attention scores of $[CLS]$ token with MF. In $w$ We can see the tokens $5$ and $may$ have higher attention scores and the tokens of 'loan' and 'contract' feature have uniformly low attention scores. That shows the ability of our method in selecting features, which addresses the problem of the neural-based method sensitive to the informative features~\cite{grinsztajn2022tree}.

\subsection{Effects on modeling feature}
To demonstrate that our method can learn contextual representations from textual features, we visualize the similarity between features. As we all know that a good representation of tabular data is sensitive and meaningful to the change of feature value. For example, two data point has very similar feature values except one feature should have different representation that reflects the difference between these feature value. For example, we enumerate the value of "job" and get the embedding of $[CLS]$ token with one randomly example from BM dataset.  We use the Euclidean distance between each pair of $[CLS]$ embeddings to measure the similarity.  As show in Fig.~\ref{fig:fig6_1}, the feature with lower importance, the most unsimilar "job" change from "unemployed"-"housemaid" to "admin"-"retired". Interesting, as show in Fig.~\ref{fig:fig6_2},  the month of "last contact" has forms three clusters: $[1,4],[5,8], [9,12]$. That shows the effectiveness of PTab in learning proper contextual representations for tabular data.





\section{Conclusion and futher works}
To model tabular data in textual modality, we propose a novel framework, the Pre-trained language model to model Tabular data (PTab). It enhances the representation of tabular data with semantic information and makes the training on mixed datasets feasible. The experiments shows that PTab outperform the-state-of-the-art baslines  on both supervised and semi-supervised settings in terms of average AUC scores. The visulizations analysis show that PTab can effectively select features by attention mechanism and learn contextual representation from textualized tabular data.

In the future, we can improve self-supervised learning in PTab. We have empirically tested the different mask ratios in MF stage and got results with an insignificant difference. The way to effectively learn representation under a self-supervised setting still needs to be explored. Besides, the precise representation of numbers is hard in textual modality. The numbers have often been tokenized into multi-tokens and that may lead to major affection for the learning of contextual representation. But this direction is beyond the topic of this paper.

\section{Acknowledgements}
This work was supported by the National Key R\&D Program of China (2020AAA0105200).
\bibliography{aaai23}

\end{document}